\documentclass{article}

  \usepackage[utf8]{inputenc} 
  \usepackage[T1]{fontenc}    
  \usepackage{url}            
  \usepackage{booktabs}       
  \usepackage{amsfonts}       
  \usepackage{nicefrac}       
  \usepackage{microtype}      
  \usepackage{verbatim}
  \usepackage{booktabs} 

  \usepackage{epsfig}
  \usepackage{graphicx}
  \usepackage{epstopdf}
  \DeclareGraphicsExtensions{.eps}
  \usepackage{multicol}
  \usepackage{pdflscape}
  \usepackage{afterpage}

  \usepackage{amssymb}
  \usepackage{amsmath}
  \usepackage{soul}
  \usepackage{xcolor}
  \usepackage{amsthm}

  \newcommand{\+}[1]{\boldsymbol{#1}}
  \newcommand{\norm}[1]{\ensuremath{\lVert{#1}\rVert}}

  \def\*#1{\mathbf{#1}}
  \def\+#1{\textit{#1}}
  \DeclareMathOperator*{\argminA}{arg\,min}
  
  \makeatletter
  
  \newcommand{\Rmnum}[1]{\expandafter\@slowromancap\romannumeral #1@}
  \makeatother

    \usepackage[accepted]{icml2017}

    \icmltitlerunning{Co-Clustering for Multitask Learning}

\begin{document}
  \twocolumn[
\icmltitle{Co-Clustering for Multitask Learning}
  



\begin{icmlauthorlist}
\icmlauthor{Keerthiram Murugesan}{cmu}
\icmlauthor{Jaime Carbonell}{cmu}
\icmlauthor{Yiming Yang}{cmu}
\end{icmlauthorlist}

\icmlaffiliation{cmu}{School of Computer Science, Carnegie Mellon University, Pittsburgh, Pennsylvania, USA. \texttt{\{kmuruges,jgc,yiming\}@cs.cmu.edu}}

\icmlcorrespondingauthor{Keerthiram Murugesan}{\texttt{kmuruges@cs.cmu.edu}}

\icmlkeywords{Multitask learning, Co-clustering, Matrix Factorization}

\vskip 0.3in
]



\printAffiliationsAndNotice{}  

  \begin{abstract}
  	This paper presents a new multitask learning framework that learns a shared representation among the tasks, incorporating both task and feature clusters. The jointly-induced clusters yield a shared latent subspace where task relationships are learned more effectively and more generally than in state-of-the-art multitask learning methods.  The proposed general framework enables the derivation of more specific or restricted state-of-the-art multitask methods. The paper also proposes a highly-scalable multitask learning algorithm, based on the new framework,  using conjugate gradient descent and generalized \+{Sylvester equations}.
  	Experimental results on synthetic and benchmark datasets show that the proposed method systematically outperforms several state-of-the-art multitask learning methods.
  \end{abstract}

  \section{Introduction}
 
  Multitask learning leverages shared structures  among the tasks to jointly build a better model for each task.  Most existing work in multitask learning focuses on how to take advantage of task similarities, either by learning the relationship between the tasks via cross-task regularization techniques \cite{zhang2014regularization,zhang2010learning,rothman2010sparse,xue2007multi} or by learning a shared feature representation across all the tasks, leveraging low-dimensional subspaces in the feature space \cite{argyriou2008convex,jalali2010dirty,liu2009multi,swirszcz2012multi}.
  Learning task relationships has been shown beneficial in (positive and negative) transfer of knowledge from  information-rich tasks to information-poor tasks \cite{zhang2014regularization}, whereas the shared feature representation has been shown to perform well when each task has a limited number of training instances (observations) compared to the total number across all tasks \cite{argyriou2008convex}. Existing research in multitask learning considers either the first approach and learns a task relationship matrix in addition to the task parameters, or relies on the latter approach and learns a shared latent feature representation from the task parameters. To the best of our knowledge, there is no prior work that utilizes both principles jointly for multitask learning. In this paper, we propose a new approach that learns a shared feature representation along with the task relationship matrix jointly to combine the advantages of both principles into a general multitask learning framework.

  Early work on latent shared representation includes \cite{zhang2005learning}, which proposes a model based on Independent Component Analysis (ICA) for learning multiple related tasks. The task parameters are assumed to be generated from independent sources. \cite{argyriou2008convex} consider sparse representations common across many learning tasks. Similar in spirit to PCA for unsupervised tasks, their approach learns a low dimensional representation of the observations \cite{ding2004k}. More recently, \cite{kumar2012learning} assume that relationships among tasks are sparse to enforce that each observed task is obtained from only a few of the latent features, and from there learn the overlapping group structure among the tasks. \cite{crammer2012learning} propose a K-means-like procedure that simultaneously clustering different tasks and learning a small pool of $m \ll T$ shared models. Specifically, each task is free to choose a model from the pool that better classifies its own data, and each model is learned from pooling together all the training data that belongs to the same cluster. \cite{barzilai2015convex} propose a similar approach that clusters the $T$ tasks into $K$ task-clusters with hard assignments. 
  
  These methods compute the factorization of the task weight matrix to learn the shared feature representation and the task structure. This matrix factorization induces the simultaneous clustering of both the tasks and the features in the $K$-dimensional latent subspace \cite{li2006relationships}. One of the major disadvantages of this assumption is that it restricts the model to define both the tasks and the features to have same number of clusters. For example, in the case of sentiment analysis, where each task belongs to a certain domain or a product category such as books, automobiles, etc., and each feature is simply a word from the vocabulary of the product reviews. Clearly, assuming both the features and the tasks have same number of clusters is an unjustified assumption, as the number of feature clusters are typically more than the number of task clusters, but the latter increase more than the former, as new products are introduced. Such a restrictive assumption may (and often does) hurt the performance of the model. 
  
  Unlike in the previous work, our proposed approach provides a flexible way to cluster both the tasks and the features. We introduce an additional degree of freedom that allows the number of task clusters to differ from the number of features clusters \cite{ding2006orthogonal,wang2011fast}. In addition, our proposed models learns both the task relationship matrix and the feature relationship matrix along with the co-clustering of both the tasks and the features \cite{gu2009co,sindhwani2009regularized}. Our proposed approach is closely related to  Output Kernel Learning (\+{OKL}) where we learn the kernel between the components of the output vector for problems such as multi-output learning, multitask learning, etc \cite{dinuzzo2011learning,sindhwani2013scalable}. The key disadvantage of \+{OKL} is that it requires the computation of kernel matrix between every pair of instances from all the tasks. This results in scalability constraint especially when the number of tasks/features is large \cite{weinberger2009feature}. Our proposed models achieve the similar effect by learning a shared feature representation common across the tasks.
  
A key challenge in factoring with the extra degree of freedom is optimizing the resulting objective function. Previous work on co-clustering for multitask learning requires strong assumptions on the task parameters. \cite{zhong2012convex}  or not scalable to large-scale applications  \cite{xu2015exploiting}.
  We propose an efficient algorithm that scales well to large-scale multitask learning  and utilizes the structure of the objective function to learn the factorized task parameters. We formulate the learning of latent variables in terms of a \+{generalized Sylvester} equation which can be efficiently solved using the conjugate gradient descent algorithm. We start from the mathematical background and then motivate our approach in Section \ref{sec:prelim}. Then
we introduce our proposed models and their learning procedures in Section \ref{sec:proposed}. Section \ref{sec:exp} reports the empirical analysis of our proposed models and shows that learning both the task clusters and the feature clusters along with the task parameters gives significant improvements compared to the state-of-the-art baselines in multitask learning. 

  \section{Background} \label{sec:prelim}


 Suppose we have $T$ tasks and $\mathcal{D}_t = \{\*X_t, Y_t\} = \{(x_{ti}, y_{ti}) : i = 1, 2, . . . , N_t\}$ is the training set for each task $t = \{1, 2, \ldots , T\}$. Let $ W_t$ represent the weight vector for a task indexed by $t$. These task weight vectors are stacked as columns of a matrix $\*W$, which is of size $P \times T$ , with $P$ being the feature dimension. Traditional multitask learning imposes additional assumptions on $
 \*W$ such as low-rank, $\ell_1$ norm, $\ell_{2,1}$ norm, etc to leverage the shared characteristics among the tasks. In this paper, we consider a similar assumption based on the factorization of the task weight matrix $\*W$.

 In factored models, we decompose the weight matrix $\*W$ as $\*\*F\*G^\top$, where $\*F$ can be interpreted as a feature cluster matrix of size $P \times K$ with $K$ feature clusters and, similarly, $\*G$ as a task cluster matrix of size $T \times K$ with $K$ task clusters. If we consider squared error losses for all the tasks, then the objective function for learning $\*F$ and $\*G$ can be given as follows:

\begin{equation}
\begin{aligned}
\argminA_{\tiny\substack{\*F \in \mathcal{R}^{P \times K} \\ \*G  \in \mathcal{R}^{T \times K}\\ \*F \in \Gamma_\mathcal{F}, \*G \in \Gamma_\mathcal{G}}}& \sum_{t \in [T]} \norm{ Y_t-\*X_t\*F G_t^\top}_2^2 +  \mathcal{P}_{\lambda_1}(\*F) + \mathcal{P}_{\lambda_2} (\*G)\\
\end{aligned}
\label{eq:genMTL}
\end{equation}
 
 In the above objective function, the latent feature representation is captured by the matrix $\*F$ and the grouping structure on the tasks is determined by the matrix $\*G$. The predictor $ W_t$ for task $t$ can then be computed from $\*F G_t^\top$, where $ G_t$ is $t^{th}$ row of matrix $\*G$. In the above objective function, $\mathcal{P}_{\lambda_1}(\*F)$ is a regularization term that penalizes the unknown matrix $\*F$ with regularization parameter $\lambda_1$. Similarly, $\mathcal{P}_{\lambda_2}(\*G)$ is a regularization term that penalizes the unknown matrix $\*G$ with regularization parameter $\lambda_2$.  $\Gamma_\mathcal{F}$ and $\Gamma_\mathcal{G}$ are their corresponding constraint spaces. Without these additional constraints on $\*F$ and $\*G$, 
 the objective function reduces to solving each task independently, since any task weight matrix from $\*F$ and $\*G$ can also be attained by $\*W$.

 Several assumptions can be enforced on these unknown factors $\*F$ and $\*G$. Below we discuss some of the previous models that make some well-known assumptions on $\*F$ and $\*G$ and can be written in terms of the above objective function.
 
 (1) \textbf{Factored Multitask Learning} (\+{FMTL}) \cite{amit2007uncovering} considers a squared {frobenius} norm on both $\*F$ and $\*G$. 
 
 \begin{equation}
\begin{aligned}
\argminA_{\tiny\substack{\*F \in \mathcal{R}^{P \times K}\\ \*G  \in \mathcal{R}^{T \times K}}}& \sum_{t \in [T]} \norm{ Y_t-\*X_t\*F G_t^\top}_2^2 + \lambda_1 \norm{\*F}_{F}^2 + \lambda_2 \norm{\*G}_{F}^2\\
\end{aligned}
\label{eq:fmtl}
\end{equation}
It can be shown that the above problem can equivalently written as the multitask learning with trace norm constraint on the task weight matrix $\*W$.
 
 (2) \textbf{Multitask Feature Learning} (\+{MTFL}) \cite{argyriou2008convex} assumes that the matrix $\*G$ learns sparse representations common across many tasks. Similar in spirit to PCA for unsupervised tasks, \textit{MTFL} learns a low dimensional representation of the observations $\*X_t$ for each task, using $\*F$ such that $\*F\*F^\top=\*I_p$.

\begin{equation}
\begin{aligned}
\argminA_{\substack{\*F \in \mathcal{R}^{P \times K},\*G  \in \mathcal{R}^{T \times K}\\ \*F\*F^\top=\*I_p}}& \sum_{t \in [T]} \norm{ Y_t-\*X_t\*F G_t^\top}_2^2 + \lambda \norm{\*G}_{2,1}^2\\
\end{aligned}
\label{eq:mtfl}
\end{equation}
where $K$ is usually set to $P$. It considers an $\ell_{2,1}$ norm on $\*G$ to force all the tasks to have a similar sparsity pattern such that the tasks select the same latent features (columns of $\*F$). It is worth noting that the Equation \ref{eq:mtfl} can be equivalently written as follows:

\begin{equation}
\begin{aligned}
\argminA_{\substack{\*W \in \mathcal{R}^{P \times T},\\ \*\Sigma \succeq 0}}& \sum_{t \in [T]} \norm{ Y_t-\*X_t W_t}_2^2 + \lambda tr({\*W^\top \*\Sigma^{-1} \*W})\\
\end{aligned}
\label{eq:mtfl2}
\end{equation}
which then can be rewritten as multitask learning with a trace norm constraint on the task weight matrix $\*W$ as before.

 (3) \textbf{Group Overlap MTL} (\+{GO-MTL}) \cite{kumar2012learning} assumes that the matrix $\*G$ is sparse to enforce that each observed task is obtained from only a few of the latent features, indexed by the non-zero pattern of the corresponding rows of the matrix $\*G$.  
 
\begin{equation}
\begin{aligned}
\argminA_{\tiny\substack{\*F \in \mathcal{R}^{P \times K}\\ \*G \in \mathcal{R}^{T \times K}}}& \sum_{t \in [T]} \norm{ Y_t-\*X_t\*F G_t^\top}_2^2 + \lambda_1 \norm{\*F}_F^2 + \lambda_2 \norm{\*G}_1^1\\
\end{aligned}
\label{eq:gomtl}
\end{equation}
 The above objective function can be compared to dictionary learning where each column of $\*F$ is considered as a dictionary atom and each row of $\*G$ is considered as their corresponding sparse codes \cite{maurer2013sparse}. 

 (4) \textbf{Multitask Learning by Clustering} (\+{CMTL}) \cite{barzilai2015convex} assumes that  the $T$ tasks  can  be clustered into $K$ task-clusters with hard assignment. For example, if the  $k$th element  of $ G_t$ is  one,  and  all  other  elements of $ G_t$ are zero, we say that task $t$ is associated with cluster $k$. 

\begin{equation}
\begin{aligned}
\argminA_{\tiny\substack{\*F \in \mathcal{R}^{P \times K},\*G \in \mathcal{R}^{T \times K} \\  G_t \in \{0,1\}^K\\ \norm{ G_t}_2=1, \forall t \in [T]}}& \sum_{t \in [T]} \norm{ Y_t-\*X_t\*F G_t^\top}_2^2 + \lambda_1 \norm{\*F}_F^2\\
\end{aligned}
\label{eq:cmtl}
\end{equation}
The constraints $ G_t \in \{0,1\}^K, \norm{ G_t}_2=1$ ensure that $\*G$ is a proper clustering matrix.
Since the above problem is computationally expensive as it involves solving a combinatorial problem, the constraint on $\*G$ is relaxed as $ G_t \in [0,1]^K$.

These four methods require the number of task clusters to be same as the  number of features clusters, which as mentioned earlier, is a restrictive assumption that may and often does hurt performance. In addition, these methods do not leverage the inherent relationship between the features (via $\*F$) and the relationship between the tasks (via $\*G$). Note that these objective functions are bi-convex problems where the optimization is convex in $\*F$ when fixing $\*G$ and vice versa. We cannot achieve globally optimal solution but one can show that algorithm reaches the locally optimal solution in a fixed number of iterations.

  \section{Proposed Approach}\label{sec:proposed}

\subsection{BiFactor MTL}
Existing models do not take into consideration both the relationship between the tasks and the relationship between the features. Here we consider a more general formulation that in addition to estimating the parameters $\*F$ and $\*G$, we learn their task relationship matrix $\*\Omega$ and the feature relationship matrix $\*\Sigma$. We call this framework \+{BiFactor} multitask learning, following the factorization of the task parameters $\*W$ into two low-rank matrices $\*F$ and $\*G$.

\begin{equation}
\begin{aligned}
\argminA_{\tiny\substack{\*F \in \mathcal{R}^{P \times K},\*G \in \mathcal{R}^{T \times K} \\ \*\Sigma \succeq 0, \*\Omega \succeq 0}}& \sum_{t \in [T]} \norm{ Y_t-\*X_t\*F G_t}_2^2 \\&+ \lambda_1 tr({\*F^\top \*\Sigma^{-1} \*F})  + \lambda_2 tr({\*G^\top \*\Omega^{-1} \*G})\\
\end{aligned}
\label{eq:fmtl1}
\end{equation}

In the above objective function, we consider $\mathcal{P}_{\lambda_1}(\*F)$ and $\mathcal{P}_{\lambda_2}(\*G)$ to learn task relationship and feature relationship matrices $\*\Sigma$ and $\*\Omega$. The motivation for these regularization terms is based on \cite{argyriou2008convex,zhang2014regularization} where they considered separately either the task relationship matrix $\*\Omega$ or the feature relationship matrix $\*\Sigma$. Note that the value of $K$ is typically set to value less than $\min(P,T)$.

It is easy to see that by setting the value of $\*G$ to $I_T$ \footnote{identity matrix of size $T \times T$ (assuming that the rank $K$ is set to $T$)} , our objective function reduces to multitask feature learning ($MTFL$) discussed in the previous section. Similarly, by setting the value of $\*F$ to $I_P$ \footnote{identity matrix of size $P \times P$ (assuming that the rank $K$ is set to $P$)} , our objective function reduces to multitask relationship learning ($MTRL$) \cite{zhang2014regularization}. If we set $\*\Omega=I_T$ and $\*\Sigma=I_P$, we obtain the factored multitask learning setting ($FMTL$) defined in Equation \ref{eq:fmtl}.  Hence the prior art can be cast as special cases of our more general formulation by imposing certain limiting restrictions.

\subsection{Optimization for \+{BiFactor} MTL}
We propose an efficient learning algorithm for solving the above objective function\+{BiFactor} MTL. Consider an alternating minimization algorithm, where we learn the shared representation $\*F$ while fixing the task structure $\*G$ and we learn the task structure $\*G$ while fixing the shared representation $\*F$. We repeat these steps until we converge to the locally optimal solution.

\+{Optimizing w.r.t $\*F$} gives an equation called generalized \textit{Sylvester equation} of the form $AQB^{\top} + CQD^{\top} = E$ for the unknown $Q$. We will show in the next section on how to solve these linear equation efficiently. From the objective function, we have:
\begin{equation} \label{eq:bcf}
\sum_t (\*X_t^\top \*X_t) \*F ( G_t^\top  G_t) + \lambda_1 \*\Sigma^{-1} F = \sum_t\*X_t^\top  Y_t  G_t
\end{equation}

\+{Optimizing w.r.t $\*G$} for squared error loss results in the similar linear equation:

\begin{equation}\label{eq:bcg}
(\*F^\top\*X_t^\top \*X_t\*F)  G_t + \lambda_1 \*\Omega^{-1}\*G = \*F^\top\*X_t^\top  Y_t
\end{equation}

\+{Optimizing w.r.t $\*\Omega$ and $\*\Sigma$}:
The optimization of the above function w.r.t $\*\Omega$ and $\*\Sigma$ while fixing the other unknowns can be learned easily with the following closed-form solutions \cite{zhang2014regularization}:

\begin{multicols}{2}
\begin{equation*}
\begin{aligned} \label{eq:bcf1}
\*\Omega=\frac{(\*G\*G^{\top})^{\frac{1}{2}}}{tr((\*G\*G^{\top})^{\frac{1}{2}})}
\end{aligned}
\end{equation*}

\begin{equation*}
\begin{aligned} \label{eq:bcf2}
\*\Sigma=\frac{(\*F \*F^{\top})^{\frac{1}{2}}}{tr((\*F\*F^{\top})^{\frac{1}{2}})}
\end{aligned}
\end{equation*}
\end{multicols}

\subsection{TriFactor MTL}
As mentioned earlier, one of the restrictions in \+{BiFactor} MTL and factored models is that both the number of feature clusters and task clusters should be set to $K$. This poses a serious model restriction, by assuming both the latent task and feature representation live in a same subspace. Such assumption can significantly hinder the flexibility of the model search space and we address this problem with a modification to our previous framework.

 Following the previous work in matrix tri-factorization, we introduce an additional factor $\*S$ such that we write $\*W$ as $\*F\*S\*G^\top$ where $\*F$ is a feature cluster matrix of size $P \times K_1$ with $K_1$ feature clusters and $\*G$ is a task cluster matrix of size $T \times K_2$ with $K_2$ task clusters and $\*S$ is the matrix that maps feature clusters to task clusters. With this representation, latent features lie in a $K_1$ dimensional subspace and the latent tasks lie in a $K_2$ dimensional subspace.

\begin{equation}
\begin{aligned}
\argminA_{\tiny\substack{\*F \in \mathcal{R}^{P \times K_1}, \\ \*G \in \mathcal{R}^{T \times K_2}, \\ \*S \in \mathcal{R}^{K_1 \times K_2} \\ \*\Sigma \succeq 0, \*\Omega \succeq 0}}& \sum_{t \in [T]} \norm{ Y_t-\*X_t\*F\*S G_t^\top}_2^2 \\ &+ \lambda_1 tr({\*F^\top \*\Sigma^{-1} \*F}) + \lambda_2 tr({\*G^\top \*\Omega^{-1} \*G})\\
\end{aligned}
\label{eq:fmtl2}
\end{equation}

The cluster mapping matrix $\*S$ introduces an additional degree of freedom in the factored models and addresses the realistic assumptions encountered in many applications. Note that we do not consider any regularization on $\*S$ in this paper, but one may impose additional constraint on $\*S$ such as $\ell_1$ (sparse penalty), $\ell_2^2$ (ridge penalty), non-negative constraints, etc, to further improve performance.

\subsection{Optimization for \+{TriFactor} MTL}
We introduce an efficient learning algorithm for solving \+{TriFactor} MTL, similar to the optimization procedure for \+{BiFactor} MTL. As before, we consider an alternating minimization algorithm, where we learn the shared representation $\*F$ while fixing the $\*G$ and $\*S$, we learn the task structure $\*G$ while fixing the $\*F$ and $\*S$ and we learn the cluster mapping matrix $\*S$, by fixing $\*F$ and $\*G$. We repeat these steps until we converge to a locally optimal solution.

\+{Optimizing w.r.t $\*F$} gives a generalized \textit{Sylvester equation} as before.
\begin{equation}\label{eq:tcf}
\sum_t (\*X_t^\top \*X_t) \*F (\*S G_t^\top  G_t\*S^\top) + \lambda_1 \*\Sigma^{-1}\*F = \sum_t\*X_t^\top  Y_t  G_t \*S^{\top}
\end{equation}

\+{Optimizing w.r.t $\*G$} gives the following linear equation:
\begin{equation}\label{eq:tcg}
(\*S^\top\*F^\top\*X_t^\top \*X_t\*F\*S)  G_t + \lambda_1 \*\Omega^{-1}\*G = \*S^\top \*F^\top\*X_t^\top  Y_t
\end{equation}
for all $t \in [T]$. 

\+{Optimizing w.r.t $\*S$}:
Solving for $\*S$ results in the following equation:
\begin{equation}\label{eq:tcs}
\sum_t (\*F^\top \*X_t^\top \*X_t \*F)\*S( G_t^\top G_t)=\sum_t \*F^\top\*X_t^\top Y_t  G_t
\end{equation}

\+{Optimizing w.r.t $\*\Omega$ and $\*\Sigma$}:
The optimization of the above function w.r.t $\*\Omega$ and $\*\Sigma$ while fixing the other unknowns can be learned as in \textit{BiFactor}MTL.  Note that one may consider $\ell_1$ regularization on $\*\Omega$ and $\*\Sigma$ to learn the sparse relationship between the tasks and the features \cite{zhang2010learning}.  

%
%

\subsection{Solving the Generalized Sylvester Equations}
We give some details on how to solve the generalized  \textit{Sylvester equations} (\ref{eq:bcf},\ref{eq:bcg},\ref{eq:tcf},\ref{eq:tcg},\ref{eq:tcs}) encountered in \textit{BiFactor} and \+{TriFactor} MTL optimization steps.  The generalized \textit{Sylvester equation} of the form $AQB^{\top} + CQD^{\top} = E$ has a unique solution $Q$ under certain regularity conditions which can be exactly obtained by an extended version of the classical Bartels-Stewart method whose complexity is $\mathcal{O}((p+q)^3)$ for $p \times q$-matrix variable $Q$, compared to the naive matrix inversion which requires $\mathcal{O}(p^3q^3)$.

 Alternatively one can solve the linear equation using the properties of the Kronecker product:  $(B^{\top} \otimes  A) vec(Q) + (D^{\top} \otimes C) vec(Q) = vec(E)$ where $\otimes$ is the Kronecker product and $vec(.)$ vectorizes $Q$ in a column oriented way. Below, we show the alternative form for \textit{TriFactor} MTL equations:
{\footnotesize
\begin{equation}
\begin{aligned}\label{eq:tcf3}
\sum_t ((\*S G_t^\top  G_t\*S^\top) \otimes (\*X_t^\top \*X_t)) vec(\*F)  &+ \lambda_1 (\*I_{K_1} \otimes \*\Sigma^{-1}) vec(\*F) \\&= vec\big(\sum_t\*X_t^\top  Y_t  G_t \*S^{\top}\big)
\end{aligned}
\end{equation}

\begin{equation}
\begin{aligned}\label{eq:tcg3}
diag(\*S^\top\*F^\top\*X_t^\top \*X_t\*F\*S)_{t=1}^T vec(\*G) &+ (\*I_{K_2} \otimes \*\Omega^{-1} )vec(\*G) \\&= vec([\*S^\top \*F^\top\*X_t^\top  Y_t]_{t=1}^T)
\end{aligned}
\end{equation}

\begin{equation}
\begin{aligned}\label{eq:tcs3}
\sum_t (( G_t^\top G_t) \otimes (\*F^\top \*X_t^\top \*X_t \*F)) vec(S) = vec\big(\sum_t\*F^\top\*X_t^\top Y_t\big)
\end{aligned}
\end{equation}
 }
 
 We can do the same for \+{BiFactor} MTL, enabling us to use conjugate gradient descent (CG) to learn our unknown factors whose complexity depends on the condition number of the matrix $(B^{\top} \otimes  A) + (D^{\top} \otimes C)$. To optimize $\*F$, $\*G$ and $\*S$, we iteratively run conjugate gradient descent for each factor while fixing the other unknowns  until a convergence condition (tolerance $ \leq 10^{-6}$) is met. In addition, CG can exploit the solution from the previous iteration, low rank structure in the equation and the fact that the matrix vector products can be computed relatively efficiently.  From our experiments. We find that our algorithm converges fast, i.e. in a few iterations.




  \section{Experiments} \label{sec:exp}

In this section, we report on experiments on both synthetic datasets and three real world datasets to evaluate the effectiveness of our proposed MTL methods.
 We compare both our models with several state-of-the-art baselines discussed in Section \ref{sec:prelim}. We include the results for {Shared Multitask learning} (\textit{SHAMO}) \cite{crammer2012learning}, which uses a K-means like procedure that simultaneously clusters different tasks using a small pool of $m \ll T$ shared model. Following \cite{barzilai2015convex}, we use gradient-projection algorithm to optimize the dual of the objective function (Equation \ref{eq:cmtl}). In addition, we compare our results with {Single-task learning} (\textit{STL}), which learns a single model by pooling together the data from all the tasks and {Independent task learning} (\textit{ITL}) which learns each task independently. 


\begin{table*}[ht]
\centering
\caption{Performance results (RMSE) on synthetic datasets. The table reports the mean and standard errors over $5$ random runs. The best model and the statistically competitive models (by paired \+{t-test} with $\alpha=0.05$) are shown in boldface.}
\label{tab:syn1}
\begin{tabular}{l|l|l|l|l|l|}\hline
\multicolumn{1}{|l|}{\textit{\textbf{Model}}} & \textbf{syn1}        & \textbf{syn2}        & \textbf{syn3}        & \textbf{syn4}        & \textbf{syn5}        \\ \hline
\multicolumn{1}{|l|}{\textit{STL}}            & 4.79 (0.04)          & 5.71 (0.05)          & 5.5 (0.04)           & 4.02 (0.02)          & 5.72 (0.06)          \\ \hline
\multicolumn{1}{|l|}{\textit{ITL}}            & 1.98 (0.08)          & 2.10 (0.09)          & 2.01 (0.06)          & 1.95 (0.06)          & 2.14 (0.07)          \\ \hline
\multicolumn{1}{|l|}{\textit{SHAMO}}          & 3.63 (0.22)          & 4.37 (0.27)          & 3.56 (0.23)          & 2.76 (0.13)          & 4.27 (0.31)          \\ \hline
\multicolumn{1}{|l|}{\textit{MTFL}}           & 1.91 (0.08)          & 1.95 (0.07)          & 1.64 (0.06)          & 1.47 (0.05)          & 1.91 (0.07)          \\ \hline
\multicolumn{1}{|l|}{\textit{GO-MTL}}         & 1.84 (0.10)          & 1.90 (0.08)          & 1.72 (0.04)          & 1.63 (0.06)          & 1.84 (0.06)          \\ \hline\hline
\multicolumn{1}{|l|}{\textit{BiFactorMTL}}    & 1.85 (0.08)          & \textbf{1.85 (0.06)} & 1.68 (0.08)          & 1.37 (0.08)          & 1.74 (0.07)          \\ \hline
\multicolumn{1}{|l|}{\textit{TriFactorMTL}}   & \textbf{1.78 (0.08)} & \textbf{1.83 (0.07)} & \textbf{1.46 (0.05)} & \textbf{1.31 (0.02)} & \textbf{1.68 (0.10)} \\ \hline
\end{tabular}
\end{table*}

The parameters for the proposed formulations and several state-of-the-art baselines are chosen from $3$-fold cross validation. We fix the value of $\lambda_1$ to $0.1$ in order to reduce the search space. The value for $\lambda_2$ is chosen from the search grid $\{10^{-3},10^{-2}, \ldots,10^{2},10^{3}\}$. The value for $K$, $K_1$ and $K_2$ are chosen from $\{2,3,5,7,9,10,15\}$. We evaluate the models using Root Mean Squared Error (\textit{RMSE}) for the regression tasks and using $F$-measure for the classification tasks. For our experiments, we consider the squared error loss for each task. We repeat all our experiments  $5$ times to compensate for statistical variability. The best model and the statistically competitive models (by paired \+{t-test} with $\alpha=0.05$) are shown in boldface. 

\subsection{Synthetic Data}
We evaluate our models on five synthetic datasets based on the assumptions considered in both the baselines and the proposed methods. We generate $100$ examples from $\*X_t \sim \mathcal{N}(0,\*I_P)$ with $P=20$ for each task $t$. All the datasets consist of $30$ tasks with $25$ training examples per task. Each task is constructed using $Y_t = \*X_t W_t +\mathcal{N}(0,1)$. The task parameters for each synthetic dataset is generated as follows:

\begin{enumerate}
     \item \textbf{\textit{syn1}} dataset consists of $3$ groups of tasks with $10$ tasks in each group without any overlap.  We generate $K=15$ latent features from $\*F \sim \mathcal{N}(0,1)$ and each $W_t$ is constructed from linearly combining $5$ latent features from $\*F$. 
     \item \textbf{\textit{syn2}} dataset is generated with $3$ overlapping groups of tasks. As before, we generate $K=15$ latent features from $\*F \sim \mathcal{N}(0,1)$ but tasks in group $1$ are constructed from features $1-7$, tasks in group $2$ are constructed from features $4-12$ and the tasks in group $3$ are constructed from features $9-15$.
     \item \textbf{\textit{syn3}} dataset simulates the \+{BiFactor} MTL. We randomly generate task covariance matrix $\*\Omega$ and feature covariance matrix $\*\Sigma$. We sample $\*F \sim \mathcal{N}(0,\*\Sigma)$ and $\*G \sim \mathcal{N}(0,\*\Omega)$ and compute $\*W=\*F\*G^\top$.
     \item \textbf{\textit{syn4}} dataset simulates the \+{TriFactor} MTL. We randomly generate task covariance matrix $\*\Omega$ and feature covariance matrix $\*\Sigma$. We sample $\*F \sim \mathcal{N}(0,\*\Sigma)$, $\*G \sim \mathcal{N}(0,\*\Omega)$ and $\*S \sim \mathcal{U}(0,1)$. We compute the task weight matrix by $\*W=\*F\*S\*G^\top$.
     \item \textbf{\textit{syn5}} dataset simulates the experiment with task weight matrix drawn from a matrix normal distribution \cite{zhang2010learning}. We randomly generate task covariance matrix $\*\Omega$ and feature covariance matrix $\*\Sigma$. We sample $vec(\*W) \sim \mathcal{N}(0,\*\Sigma \otimes \*\Omega)$. 
\end{enumerate}

\begin{table*}[ht]
\centering
\caption{Performance results (RMSE) on school datasets. The table reports the mean and standard errors over $5$ random runs. }
\label{tab:school}
\begin{tabular}{|l|l|l|l|l|l||l|l|}
\hline
\textit{\textbf{Models}} & \textit{\textbf{STL}} & \textit{\textbf{ITL}} & \textit{\textbf{SHAMO}} & \textit{\textbf{MTFL}} & \textit{\textbf{GO-MTL}} & \textit{\textbf{BiFactor}} & \textit{\textbf{TriFactor}} \\ \hline
\textit{20\%}            & 12.19 {\tiny(0.03)}          & 12.00 {\tiny(0.04)}          & 11.91 {\tiny(0.05)}            & 11.25 {\tiny(0.05)}           & 11.15 {\tiny(0.05)}             & 10.68 {\tiny(0.08)}                  & \textbf{10.54 {\tiny(0.09)}}          \\ \hline
\textit{30\%}            & 12.09 {\tiny(0.07)}          & 12.01 {\tiny(0.05)}          & 10.92 {\tiny(0.05)}            & 10.85 {\tiny(0.02)}           & 10.53 {\tiny(0.10)}             & 10.38 {\tiny(0.11)}                  & \textbf{10.22 {\tiny(0.08)}}          \\ \hline
\textit{40\%}            & 12.00 {\tiny(0.10)}          & 11.88 {\tiny(0.06)}          & 11.82 {\tiny(0.06)}            & 10.61 {\tiny(0.06)}           & 10.31 {\tiny(0.14)}             & 10.20 {\tiny(0.13)}                  & \textbf{10.12 {\tiny(0.10)}}          \\ \hline
\end{tabular}
\end{table*}

 We compare the proposed methods \textit{BiFactor} MTL and \textit{TriFactor} MTL against the baselines. We can see in Table \ref{tab:syn1} that \textit{BiFactor} and \textit{TriFactor} MTL outperforms all the baselines in all the synthetic datasets. \+{STL} performs the worst since it combines the data from all the tasks. We can see that the \+{SHAMO} performs better than \+{STL} but worse than \+{ITL} which shows that learning these tasks separately is beneficial than combining them to learn a fewer models. 
 
 As mentioned earlier, since \+{MTFL} is similar to \+{FMTL} in Equation \ref{eq:fmtl}, we can see how the results of \textit{BiFactor} MTL improve when it learns both the task relationship matrix and the feature relationship matrix.
 Note that the \textbf{\textit{syn1}} and \textbf{\textit{syn2}} datasets are based on assumptions in \+{GO-MTL}, hence, it performs better than the other baselines. \textit{BiFactor} MTL and \textit{TriFactor} MTL models are equally competent with \+{GO-MTL} which shows that our proposed methods can easily adapt to these assumptions. Synthetic datasets \textbf{\textit{syn3}}, \textbf{\textit{syn4}} and \textbf{\textit{syn5}} are generated with both the task covariance matrix and the feature covariance matrix. Since both \textit{BiFactor} MTL and \textit{TriFactor} MTL learns task and feature relationship matrix along with the task weight parameters, they performs significantly better than other baselines. 
 
\subsection{Exam Score Prediction}
We evaluate the proposed methods on examination score prediction data, a benchmark dataset in multitask regression reported in several previous articles \cite{argyriou2008convex,kumar2012learning,zhang2014regularization} \footnote{\url{http://ttic.uchicago.edu/~argyriou/code/mtl_feat/school_splits.tar}}. The \textit{school} dataset consists of examination scores of $15,362$ students from $139$ schools in London. Each school is considered as a task and we need to predict exam scores for students from these $139$ schools. The feature set includes  the year of the examination, four school-specific and three student-specific attributes. We replace each categorical attribute with one binary variable for each possible attribute value, as suggested in \cite{argyriou2008convex}. This results in $26$ attributes with an additional attribute to account for the bias term. 

 Clearly, the dataset has the school and student specific feature clusters that can help in learning the shared feature representation better than the other factored baselines. In addition, there must be several task clusters in the data to account for the differences among the schools. The training and test sets are obtained by dividing examples of each task into many small datasets, by varying the size of the training data with $20\%$, $30\%$ and $40\%$, in order to evaluate the proposed methods on many tasks with limited numbers of examples.

Table \ref{tab:school} shows the experimental results for \textit{school} data.  All the factorized MTL methods outperform \+{STL} and \+{ITL}.  We can see that both \textit{TriFactor} MTL and \textit{BiFactor} MTL outperform other baselines significantly. It is interesting to see that \textit{TriFactor} MTL performs considerably well even when the tasks have limited numbers of examples.  When there is more training data, the result the advantage of \textit{TriFactor} MTL over the strongest baseline \+{GO-MTL} is reduced.

\subsection{Sentiment Analysis}
 We follow the experimental setup in \cite{crammer2012learning,barzilai2015convex} and evaluate our algorithm on product reviews from amazon\footnote{\url{http://www.cs.jhu.edu/~mdredze/datasets/sentiment}}. The dataset contains product reviews from $14$ domains such as books, dvd, etc. We consider each domain as a binary classification task.  The reviews are stemmed and  stopwords are removed from the review text. We represent each review as a bag of $5,000$ unigrams/bigrams with \+{TF-IDF} scores. We choose $1,000$ reviews from each domain and each review is associated with a rating from $\{1,2,4,5\}$. The reviews with rating $3$ is not included in this experiment as such sentiments were ambiguous and therefore cannot be reliably predicted. h
 
 We ran several experiments on this dataset to test the importance of learning shared feature representation and co-clustering of tasks and features. In Experiment \textbf{\Rmnum{1}}, we construct $14$ classification tasks with reviews labeled positive $(+)$ when rating $< 3$ and labeled negative $(-)$ when rating $< 3$. We use $240$ training examples for each task and the remaining for test set. Since all the tasks are essentially same, \+{ITL} perform better than all the other models (with an $F$-measure of $0.749$) by combining data from all the other tasks. The results for our proposed methods \textit{BiFactor} MTL ($0.722$) and \textit{TriFactor} MTL ($0.733$) are comparable to that of \+{ITL}. See supplementary material for the results of Experiment \textbf{\Rmnum{1}}.

 For Experiment \textbf{\Rmnum{2}}, we split each domain into \+{two} equal sets, from which we create two prediction tasks based on the two different thresholds: whether the rating for the reviews is $5$ or not and whether the rating for the reviews is $1$ or not. Obviously, combining all the tasks together will not help in this setting.  Experiments \textbf{\Rmnum{3}} and  \textbf{\Rmnum{4}} are similar to Experiment \textbf{\Rmnum{2}}, except that each task is further divided  into $2$ or $3$ sub-tasks.
 
 Experiment \textbf{\Rmnum{5}} splits each domain into \+{three} equal sets to construct three prediction tasks based on three different thresholds: whether the rating for the reviews is $5$ or not, whether the rating for the reviews is $< 3$ or not  and whether the rating for the reviews is $1$ or not. This setting captures the reviews with different levels of sentiments. As before, we build the dataset for Experiments \textbf{\Rmnum{6}} and  \textbf{\Rmnum{7}} by further dividing the three prediction tasks from Experiment \textbf{\Rmnum{5}} into $2$ or $3$ sub-tasks.
 
 The results from our experiments are reported in Table \ref{tab:sent}. The first four rows in the table show the number of tasks in each experiment, number of thresholds considered for the ratings, number of splits constructed  from each domain and the total number of training examples in each task. The general trend is that factorized models performs significantly better than the other baselines. Since \+{MTFL}, \+{BiFactor}MTL and \+{TriFactor}MTL learn feature relationship matrix $\*\Sigma$ in addition to the task parameter, they achieve better results than \+{CMTL}, which considers only the task clusters. 

We notice that as we increase the number of tasks, the gap between the performances of \+{TriFactor}MTL and \+{BiFactor}MTL (and \+{GO-MTL}) widens, since the assumption that the the number of feature and task clusters $K$ should be same is clearly violated. On the other hand, \+{TriFactor}MTL learns with a different number of feature and task clusters $(K_1,K_2)$ and, hence achieves a better performance than all the other methods considered in these experiments.

\begin{table*}[ht]
\centering
\caption{Performance results (F-measure) for various experiments on sentiment detection. The table reports the mean and standard errors over $5$ random runs. }
\label{tab:sent}
\begin{tabular}{|c|c|c|c||c|c|c|}
\hline
\textit{\textbf{Data}}                                                  & \textbf{\Rmnum{2}}              & \textbf{\Rmnum{3}}              & \textbf{\Rmnum{4}}              & \textbf{\Rmnum{5}}              & \textbf{\Rmnum{6}}              & \textbf{\Rmnum{7}}              \\ \hline
\textit{Tasks}                                                          & 28                                & 56                                & 84                                & 42                                & 86                                & 126                               \\ \hline
\textit{\begin{tabular}[c]{@{}c@{}}Thresholds \\ (Splits)\end{tabular}} & 2 (2)                             & 2 (4)                             & 2 (6)                             & 3 (3)                             & 3 (6)                             & 3 (9)                             \\ \hline
\textit{Train Size}                                                     & 120                               & 60                                & 40                                & 80                                & 40                                & 26                                \\ \hline\hline
\textit{STL}                                                            & 0.429 {\tiny(0.002)}                      & 0.432 {\tiny(0.001)}                      & 0.429 {\tiny(0.002)}                      & 0.400 {\tiny(0.002)}                      & 0.399 {\tiny(0.003)}                      & 0.397 {\tiny(0.001)}                      \\ \hline
\textit{ITL}                                                            & 0.433 {\tiny(0.001)}                      & 0.440 {\tiny(0.002)}                      & 0.431 {\tiny(0.001)}                      & 0.499 {\tiny(0.001)}                      & 0.486 {\tiny(0.002)}                      & 0.479 {\tiny(0.001)}                      \\ \hline
\textit{SHAMO}                                                          & 0.423 {\tiny(0.002)}                      & 0.437 {\tiny(0.006)}                      & 0.429 {\tiny(0.002)}                      & 0.498 {\tiny(0.006)}                      & 0.460 {\tiny(0.002)}                      & 0.496 {\tiny(0.013)}                      \\ \hline
\textit{CMTL}                                                           & 0.557 {\tiny(0.016)}                      & 0.436 {\tiny(0.007)}                      & 0.429 {\tiny(0.004)}                      & 0.508 {\tiny(0.002)}                      & 0.486 {\tiny(0.002)}                      & 0.476 {\tiny(0.002)}                      \\ \hline
\textit{MTFL}                                                           & 0.482 {\tiny(0.004)}                      & 0.473 {\tiny(0.002)}                      & 0.432 {\tiny(0.007)}                      & 0.522 {\tiny(0.002)}                      & 0.487 {\tiny(0.003)}                      & 0.481 {\tiny(0.002)}                      \\ \hline
{GO-MTL}                                            & 0.582 {\tiny(0.012)} & 0.526 {\tiny(0.013)} & 0.516 {\tiny(0.007)} & 0.587 {\tiny(0.004)} & 0.540 {\tiny(0.005)} & 0.539 {\tiny(0.008)} \\ \hline \hline
\textit{BiFactor}                                                       & 0.611 {\tiny(0.018)}                      & 0.561 {\tiny(0.013)}                      & \textbf{0.598 {\tiny(0.002)}}             & \textbf{0.643 {\tiny(0.013)}}             & 0.578 {\tiny(0.020)}                      & 0.574 {\tiny(0.052)}                      \\ \hline
\textit{TriFactor}                                                      & \textbf{0.627 {\tiny(0.008)}}             & \textbf{0.588 {\tiny(0.006)}}             & \textbf{0.603 {\tiny(0.012)}}             & \textbf{0.655 {\tiny(0.013)}}             & \textbf{0.606 {\tiny(0.020)}}             & \textbf{0.632 {\tiny(0.029)}}             \\ \hline
\end{tabular}
\end{table*}

\begin{table*}[!ht]
\centering
\caption{Performance results (F-measure) on 20\+{Newsgroups} dataset. The table reports the mean and standard errors over $5$ random runs. }
\label{tab:20ng}
\begin{tabular}{|l|l|l|l|l|l|}
\hline
\textit{\textbf{Models}} & \textbf{Task 1}      & \textbf{Task 2}      & \textbf{Task 3}      & \textbf{Task 4}      & \textbf{Task 5}      \\ \hline
\textit{GO-MTL}          & 0.42 (0.09)          & 0.57 (0.06)          & 0.42 (0.04)          & 0.47 (0.06)          & 0.40 (0.03)          \\ \hline
\textit{BiFactorMTL}     & 0.42 (0.09)          & 0.60 (0.05)          & 0.41 (0.04)          & 0.49 (0.03)          & 0.36 (0.01)          \\ \hline
\textit{TriFactorMTL}    & \textbf{0.49 (0.03)} & \textbf{0.63 (0.02)} & \textbf{0.54 (0.02)} & \textbf{0.54 (0.02)} & \textbf{0.51 (0.02)} \\ \hline
\end{tabular}
\end{table*}

\subsection{Transfer Learning}

Finally, we evaluate our proposed models on 20\+{Newsgroups} dataset for transfer learning \footnote{\url{http://qwone.com/~jason/20Newsgroups/}}.
The dataset contains postings from 20  Usenet newsgroups. As before, the postings are stemmed and the stopwords are removed from the text. We represent each posting as a bag of $500$ unigrams/bigrams with \+{TF-IDF} scores. We construct $10$ tasks from the postings of the newsgroups. We randomly select a pair of newsgroup classes to build  each one-vs-one classification task. We follow the hold-out experiment suggested by \cite{raina2006constructing} for the transfer learning setup. For each of the $10$ tasks (target task), we learn $\*F$ ($\*F$ and $\*S$ in case of \+{TriFactor}MTL) from the remaining $9$ tasks (source tasks).

With $\*F$ ($\*F$ and $\*S$) known from the source tasks, we select $10\%$ of the data from the target task to learn $G_{target}$.  This experiment shows how well the learned latent feature representation from the source tasks in a $K$-dimensional subspace ($K_1$-dimensional subspace for \+{TriFactor}MTL)  adapt to the new task. We evaluate our results on the remaining data from the target task. We select \+{GO-MTL} as our baseline to compare our results. Since \+{CMTL} doesn't explicitly learn $\*F$, we did not include it in this experiment. 

Table \ref{tab:20ng} shows the results for this experiment. We report the first $5$ tasks here. See supplementary material for the performance results of all the $10$ tasks. We see that both \+{GO-MTL} and \+{BiFactor}MTL perform almost the same, since both of them learn the latent feature representation in a $K$-dimensional space. As is evident from the table,  \+{TriFactor}MTL outperforms both \+{GO-MTL} and \+{BiFactor}MTL, which shows that learning both the factors $\*F$ and $\*S$ improves information transfer from the source tasks to the target task.

  \section{Conclusions}
  
  In this paper, we proposed a novel framework for multitask learning that factors the task parameters into a shared feature representation and a task structure to learn from multiple related tasks. We formulated two approaches, motivated from recent work in multitask latent feature learning. The first (\textit{BiFactor} MTL), decomposes the task parameters $\*W$  into two low-rank matrices: latent feature representation $\*F$ and task structure $\*G$. As this approach is restrictive on the number of clusters in the latent feature and task space,  we proposed a second method (\textit{TriFactor} MTL), which introduces an additional degree of freedom to permit different clusterings in each. We developed a highly scalable and efficient learning algorithm using conjugate gradient descent and generalized Sylvester equations. Extensive empirical analysis on both synthetic and real datasets show that \textit{Trifactor} multitask learning outperforms the other state-of-the-art multitask baselines, thereby demonstrating the effectiveness of the proposed approach.

\bibliographystyle{icml2017}
\bibliography{myref}

\section*{Sensitivity Analysis}

\begin{figure*}[ht!]
	\centering
	\includegraphics[width=3in]{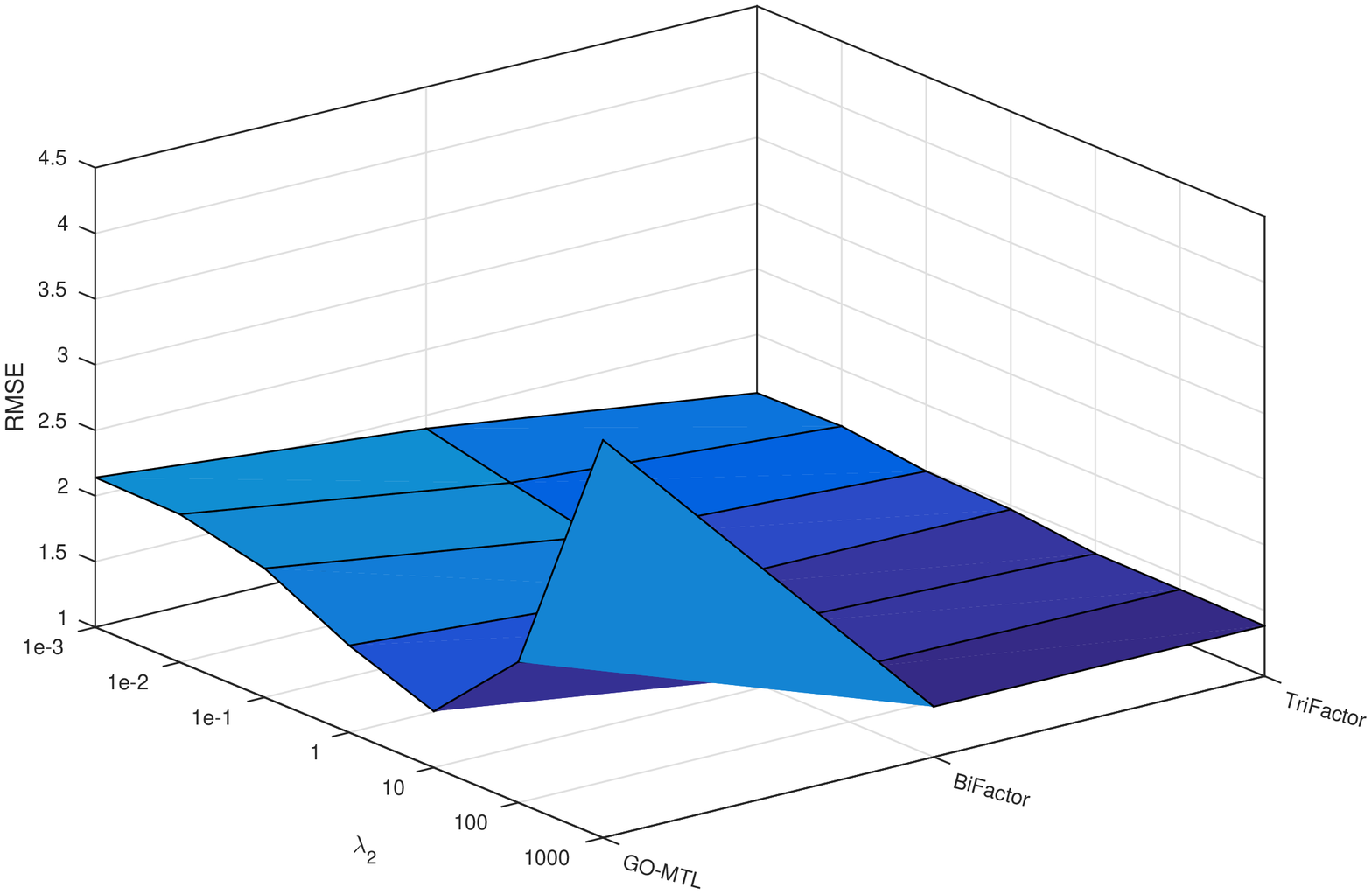}
	\hspace{0.5em}
	\includegraphics[width=3in]{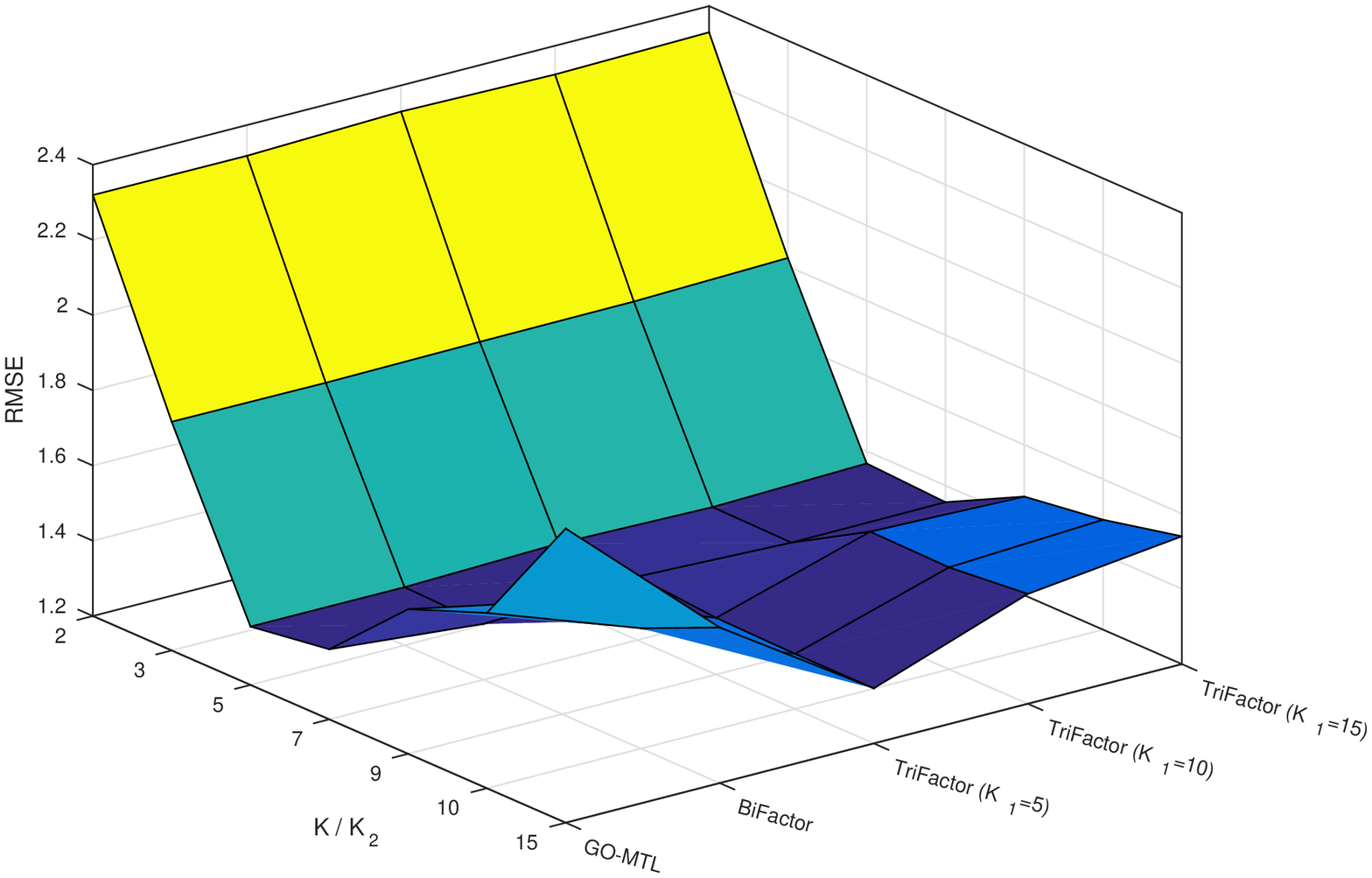}

	\includegraphics[width=3in]{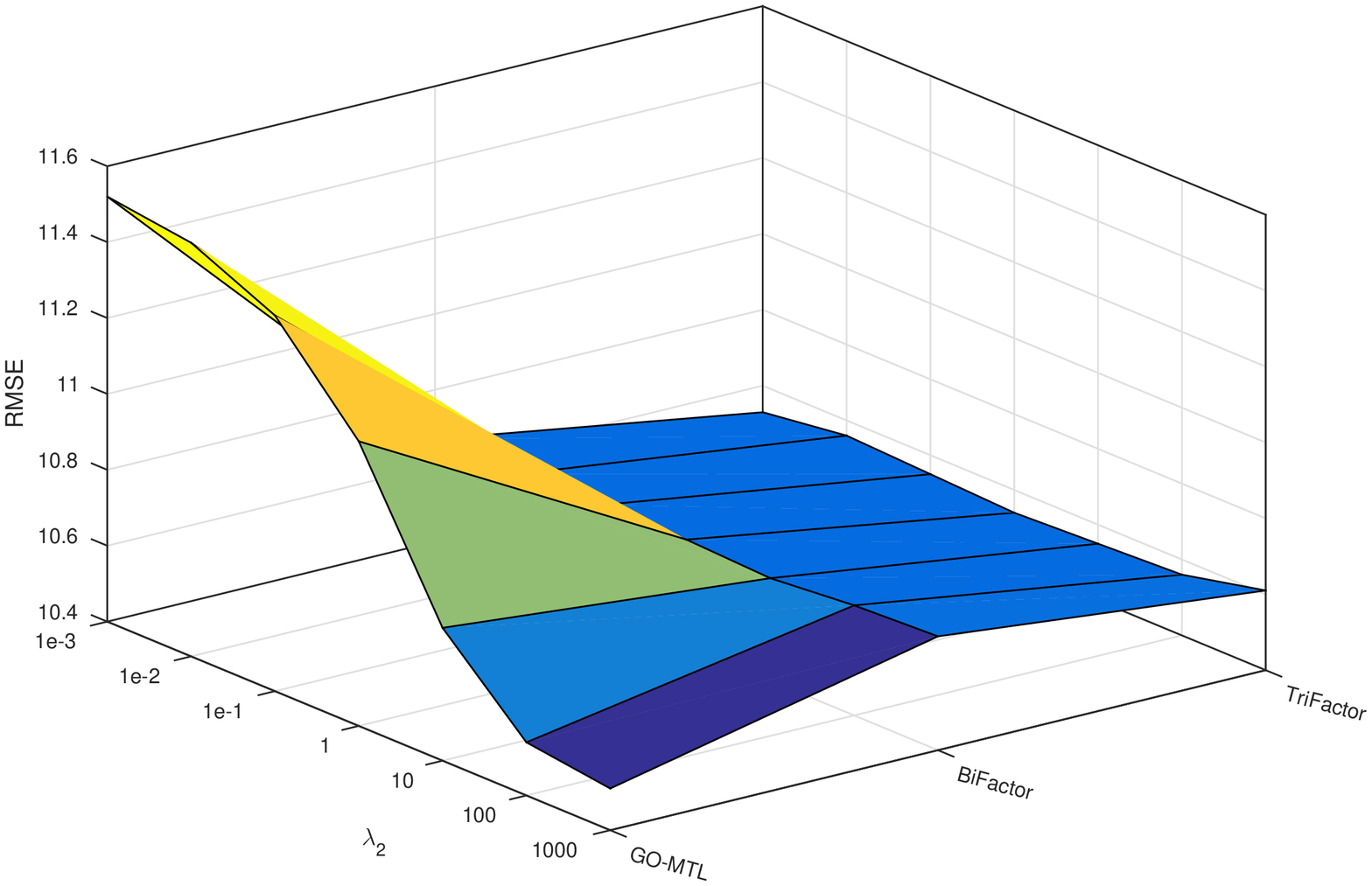}
	\hspace{0.5em}
	\includegraphics[width=3in]{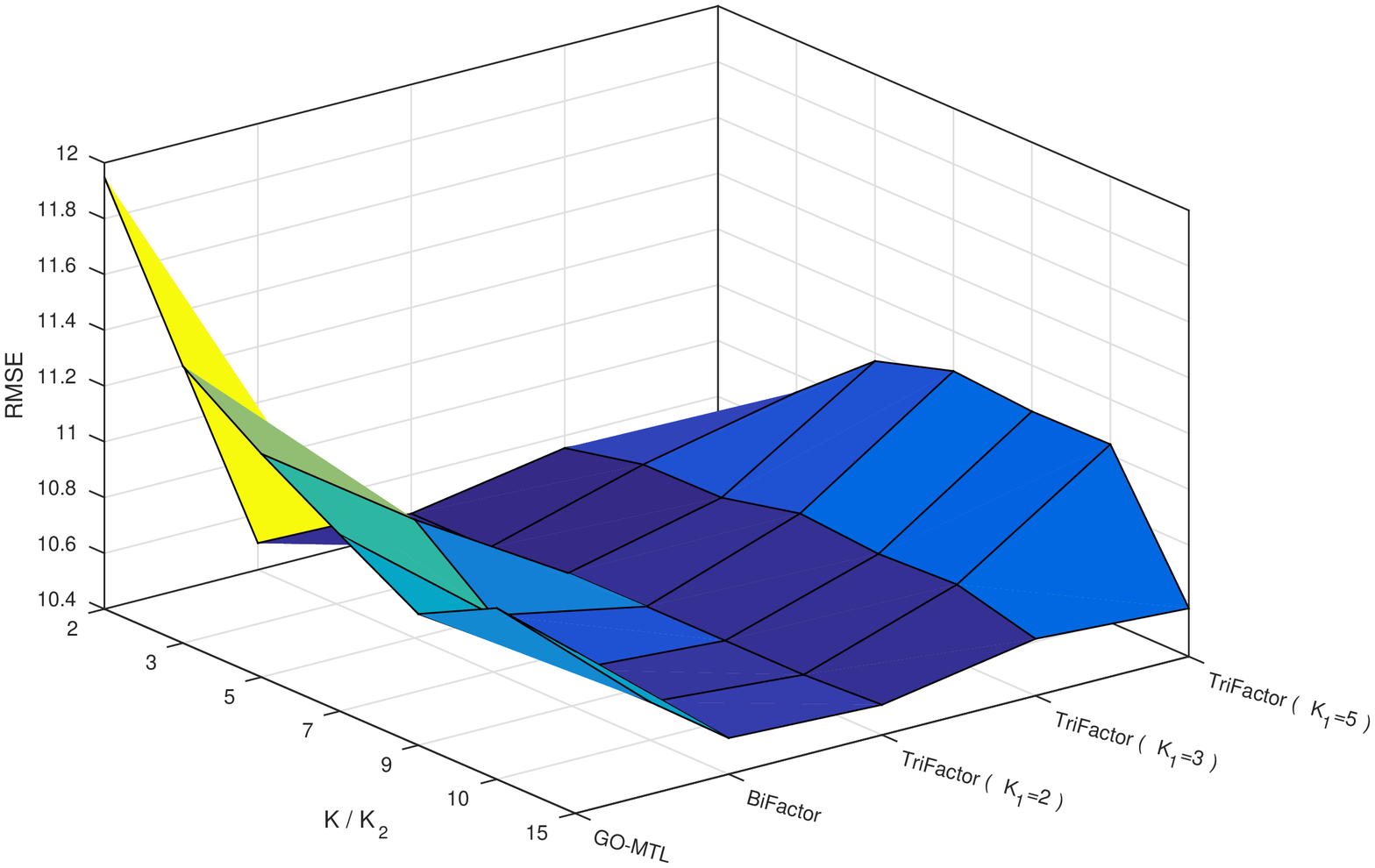}

	\includegraphics[width=3in]{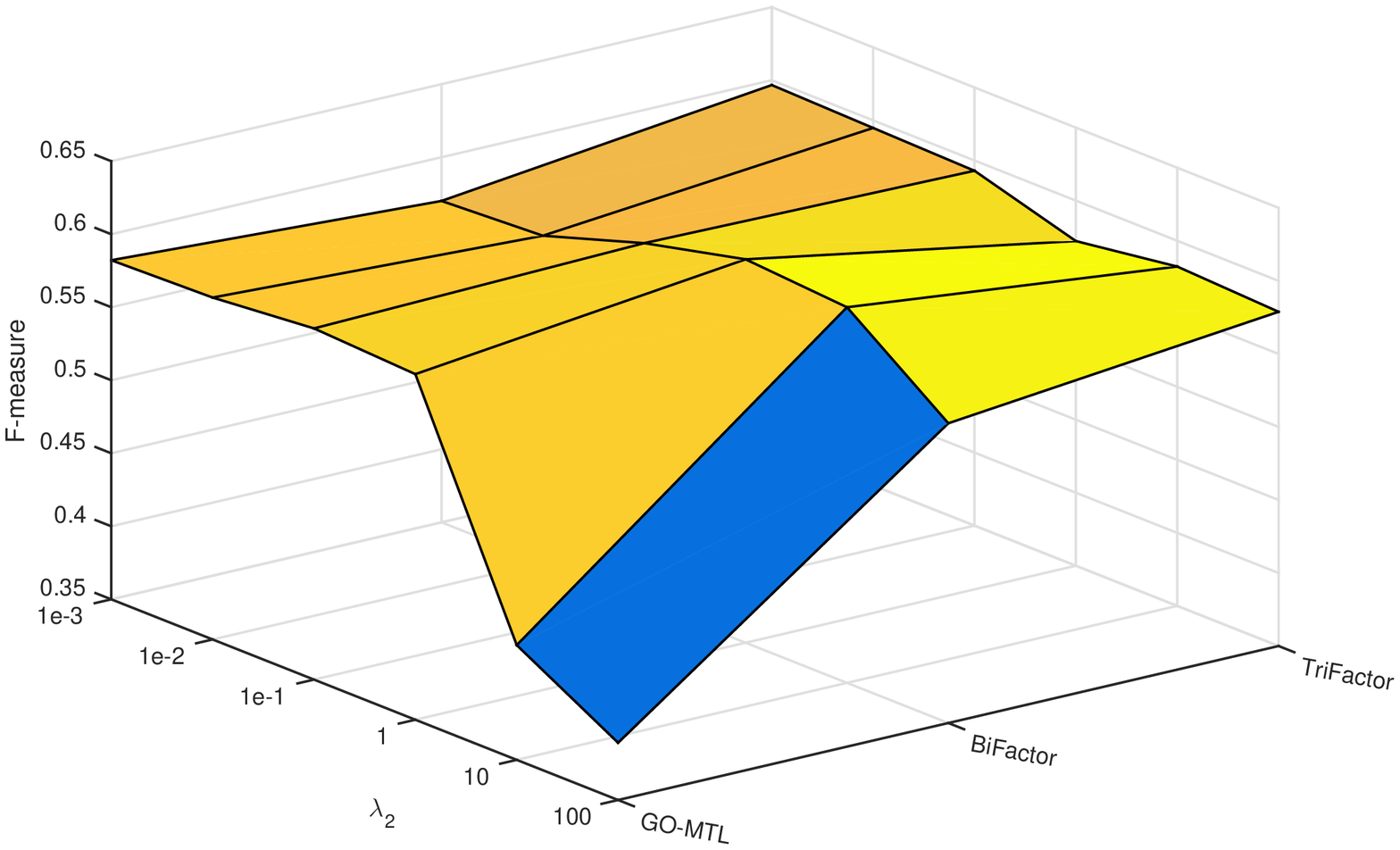}
	\hspace{0.5em}
	\includegraphics[width=3in]{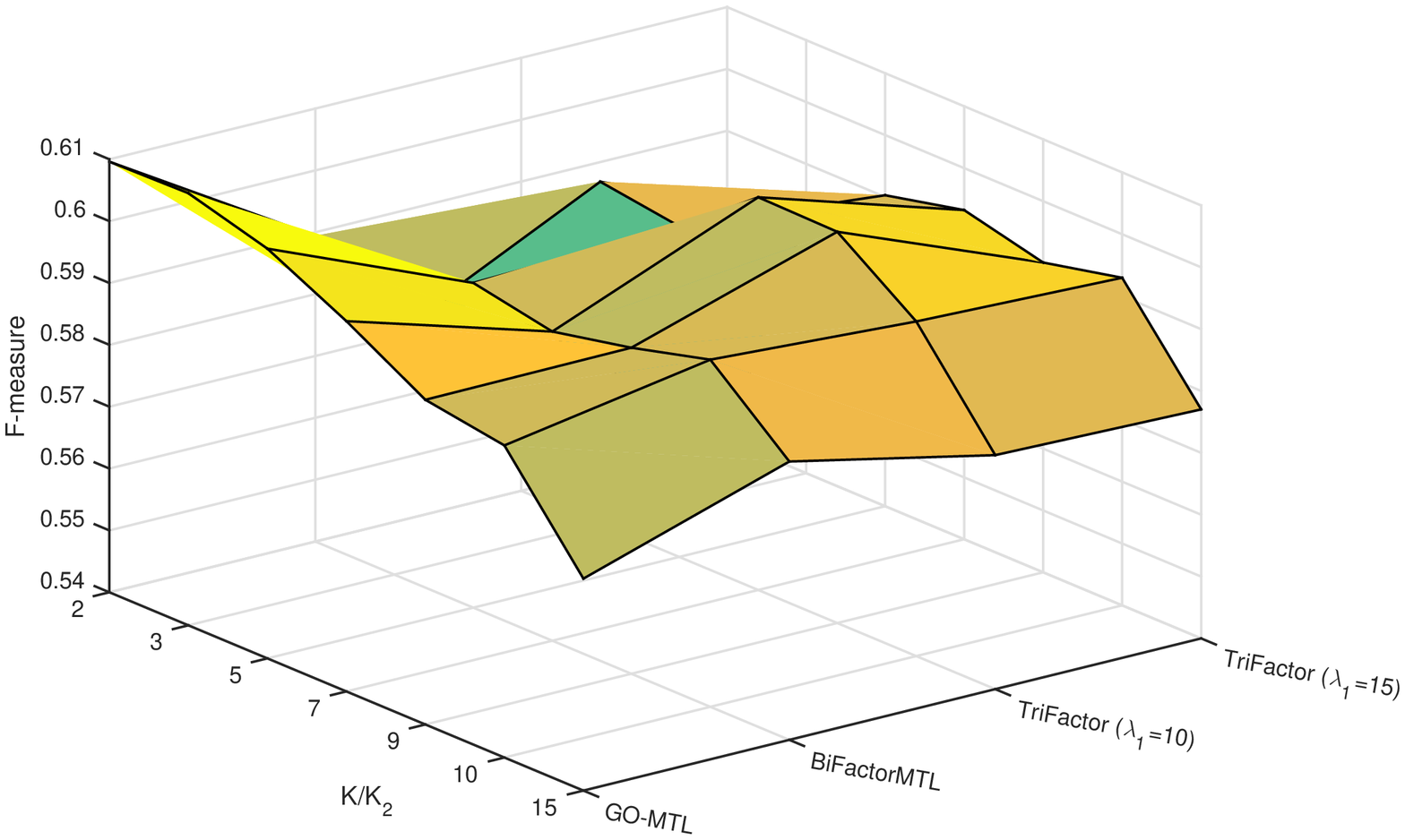}
	\caption{\textbf{Top}: Sensitivity analysis for the regularization parameter $\lambda_2$ when $\lambda_1=0.1$ and $K=2$ (left) and number of clusters $K_1$ and $K_2$ when $\lambda_1=0.1$ and $\lambda_2=0.1$ (right) calculated for \textit{syn5} dataset (RMSE).  \textbf{Middle}: Sensitivity analysis for \+{school} dataset (RMSE). \textbf{Bottom}: Sensitivity analysis for \+{sentiment} detection (F-measure).}
	\label{fig:obj}
\end{figure*}

Figure \ref{fig:obj} shows the hyper-parameter sensitivity analysis for \+{GO-MTL}, \+{BiFactor}MTL and \+{TriFactor}MTL. As before, we fix $\lambda_1=0.1$. \+{GO-MTL} and \+{BiFactor}MTL have two hyper-parameters $\lambda_2,K$ to tune and \+{TriFactor}MTL have three hyper-parameters $\lambda_2,K_1$ and $K_2$ to tune.  We can see from the plots that our proposed models yield stable results even when we change the $K, K_1$ and $K_2$. On the other hand, \+{GO-MTL} results are sensitive to the values of $\lambda_2$, regularization parameter for sparse penalty on $\*G$.

\section*{Additional Results}
Tables  \ref{tab:sentfull} and \ref{tab:transfull} show the complete experimental results for sentiment analysis and transfer learning experiments.

\begin{center}
List of one-vs-one classification tasks used in Table \ref{tab:transfull}
\begin{verbatim}
(Task 1) comp.windows.x          
vs comp.os.ms-windows.misc 
(Task 2) soc.religion.christian  
vs rec.sport.hockey        
(Task 3) misc.forsale            
vs talk.politics.guns      
(Task 4)sci.med                 
vs rec.autos               
(Task 5) comp.sys.mac.hardware   
vs talk.politics.misc      
(Task 6) sci.space               
vs alt.atheism             
(Task 7) comp.graphics           
vs comp.sys.ibm.pc.hardware
(Task 8) talk.politics.mideast   
vs sci.electronics         
(Task 9) rec.motorcycles         
vs talk.religion.misc      
(Task 10) rec.sport.baseball      
vs sci.crypt
\end{verbatim}
\end{center}

\begin{landscape}
\begin{table}[h]
\centering
\caption{Performance results (F-measure) for various experiments on sentiment detection. The table reports the mean and standard errors over $5$ random runs.}
\label{tab:sentfull}
\begin{tabular}{|c|c||c|c|c||c|c|c|}
\hline
\textit{\textbf{Data}}                                                  & \textbf{\Rmnum{1}}              & \textbf{\Rmnum{2}}              & \textbf{\Rmnum{3}}              & \textbf{\Rmnum{4}}              & \textbf{\Rmnum{5}}              & \textbf{\Rmnum{6}}              & \textbf{\Rmnum{7}}              \\ \hline
\textit{Tasks}                                                          & 14                                & 28                                & 56                                & 84                                & 42                                & 86                                & 126                               \\ \hline
\textit{\begin{tabular}[c]{@{}c@{}}Thresholds \\ (Splits)\end{tabular}} & 1 (1)                             & 2 (2)                             & 2 (4)                             & 2 (6)                             & 3 (3)                             & 3 (6)                             & 3 (9)                             \\ \hline
\textit{Train Size}                                                     & 240                               & 120                               & 60                                & 40                                & 80                                & 40                                & 26                                \\ \hline\hline
\textit{STL}                                                            & \textbf{0.749 (0.003)}             & 0.429 (0.002)                      & 0.432 (0.001)                      & 0.429 (0.002)                      & 0.400 (0.002)                      & 0.399 (0.003)                      & 0.397 (0.001)                      \\ \hline
\textit{ITL}                                                            & 0.713 (0.002)                      & 0.433 (0.001)                      & 0.440 (0.002)                      & 0.431 (0.001)                      & 0.499 (0.001)                      & 0.486 (0.002)                      & 0.479 (0.001)                      \\ \hline
\textit{SHAMO}                                                          & 0.721 (0.005)                      & 0.423 (0.002)                      & 0.437 (0.006)                      & 0.429 (0.002)                      & 0.498 (0.006)                      & 0.460 (0.002)                      & 0.496 (0.013)                      \\ \hline
\textit{CMTL}                                                           & 0.713 (0.002)                      & 0.557 (0.016)                      & 0.436 (0.007)                      & 0.429 (0.004)                      & 0.508 (0.002)                      & 0.486 (0.002)                      & 0.476 (0.002)                      \\ \hline
\textit{MTFL}                                                           & 0.711 (0.002)                      & 0.482 (0.004)                      & 0.473 (0.002)                      & 0.432 (0.007)                      & 0.522 (0.002)                      & 0.487 (0.003)                      & 0.481 (0.002)                      \\ \hline
{GO-MTL}                                            & {0.638 (0.006)} & {0.582 (0.012)} & {0.526 (0.013)} & {0.516 (0.007)} & {0.587 (0.004)} & {0.540 (0.005)} & {0.539 (0.008)} \\ \hline \hline
\textit{BiFactor}MTL                                                       & 0.722 (0.006)                      & 0.611 (0.018)                      & 0.561 (0.013)                      & \textbf{0.598 (0.002)}             & \textbf{0.643 (0.013)}             & 0.578 (0.020)                      & 0.574 (0.052)                      \\ \hline
\textit{TriFactor}MTL                                                      & 0.733 (0.006)                      & \textbf{0.627 (0.008)}             & \textbf{0.588 (0.006)}             & \textbf{0.603 (0.012)}             & \textbf{0.655 (0.013)}             & \textbf{0.606 (0.020)}             & \textbf{0.632 (0.029)}             \\ \hline

\end{tabular}
\end{table} 

\begin{table}[h!]
\centering
\caption{Performance results (F-measure) on 20\+{Newsgroups} dataset. The table reports the mean and standard errors over $5$ random runs. The best model and the statistically competitive models (by paired \+{t-test} with $\alpha=0.05$) are shown in boldface.}
\label{tab:transfull}
\begin{tabular}{|l|l|l|l|l|l|l|l|l|l|l|}
\hline
\textit{\textbf{Models}} & \textbf{Task 1}      & \textbf{Task 2}      & \textbf{Task 3}      & \textbf{Task 4}      & \textbf{Task 5}      & \textbf{Task 6}      & \textbf{Task 7}      & \textbf{Task 8} & \textbf{Task 9}      & \textbf{Task 10}     \\ \hline
\textit{GO-MTL}          & 0.42 {\tiny(0.09)}          & 0.57 {\tiny(0.06)}          & 0.42 {\tiny(0.04)}          & 0.47 {\tiny(0.06)}          & 0.40 {\tiny(0.03)}          & 0.37 {\tiny(0.02)}          & 0.35 {\tiny(0.02)}          & 0.70 {\tiny(0.01)}     & 0.38 {\tiny(0.00)}          & 0.42 {\tiny(0.05)}          \\ \hline
\textit{BiFactorMTL}     & 0.42 {\tiny(0.09)}          & 0.60 {\tiny(0.05)}          & 0.41 {\tiny(0.04)}          & 0.49 {\tiny(0.03)}          & 0.36 {\tiny(0.01)}          & 0.42 {\tiny(0.02)}          & 0.37 {\tiny(0.02)}          & 0.64 {\tiny(0.02)}     & 0.38 {\tiny(0.00)}          & 0.46 {\tiny(0.04)}          \\ \hline
\textit{TriFactorMTL}    & \textbf{0.49 {\tiny(0.03)}} & \textbf{0.63 {\tiny(0.02)}} & \textbf{0.54 {\tiny(0.02)}} & \textbf{0.54 {\tiny(0.02)}} & \textbf{0.51 {\tiny(0.02)}} & \textbf{0.67 {\tiny(0.01)}} & \textbf{0.47 {\tiny(0.02)}} & 0.66 {\tiny(0.01)}     & \textbf{0.59 {\tiny(0.03)}} & \textbf{0.62 {\tiny(0.01)}} \\ \hline
\end{tabular}
\end{table}
\end{landscape}

  \end{document}